%% file: main.tex
\documentclass{article}

\usepackage{microtype}
\usepackage{graphicx}
\usepackage{booktabs} % for professional tables
\usepackage[shortlabels]{enumitem}
\usepackage{amsmath}
\usepackage{amsfonts}
\usepackage{subfig}

\usepackage{hyperref}

\usepackage[accepted]{icml2018}
\icmltitlerunning{Behaviour Policy Estimation in Off-Policy Policy Evaluation: Calibration Matters}

\begin{document}

\twocolumn[
\icmltitle{Behaviour Policy Estimation in Off-Policy Policy Evaluation: \\ Calibration Matters}

% \icmlsetsymbol{equal}{*}

\begin{icmlauthorlist}
\icmlauthor{Aniruddh Raghu}{cam}
\icmlauthor{Omer Gottesman}{h}
\icmlauthor{Yao Liu}{s}
\icmlauthor{Matthieu Komorowski}{icl}
\icmlauthor{Aldo Faisal}{icl}
\icmlauthor{Finale Doshi-Velez}{h}
\icmlauthor{Emma Brunskill}{s}
\end{icmlauthorlist}

\icmlaffiliation{s}{Stanford University}
\icmlaffiliation{h}{Harvard University}
\icmlaffiliation{cam}{Cambridge University}
\icmlaffiliation{icl}{Imperial College London}

\icmlcorrespondingauthor{Aniruddh Raghu}{aniruddhraghu@gmail.com}

% You may provide any keywords that you
% find helpful for describing your paper; these are used to populate
% the "keywords" metadata in the PDF but will not be shown in the document
\icmlkeywords{Machine Learning, ICML}

\vskip 0.3in
]

% this must go after the closing bracket ] following \twocolumn[ ...

% This command actually creates the footnote in the first column
% listing the affiliations and the copyright notice.
% The command takes one argument, which is text to display at the start of the footnote.
% The \icmlEqualContribution command is standard text for equal contribution.
% Remove it (just {}) if you do not need this facility.

%\printAffiliationsAndNotice{}  % leave blank if no need to mention equal contribution
\printAffiliationsAndNotice{\icmlEqualContribution} % otherwise use the standard text.

\begin{abstract}
In this work, we consider the problem of estimating a behaviour policy for use in Off-Policy Policy Evaluation (OPE) when the true behaviour policy is unknown. Via a series of empirical studies, we demonstrate how accurate OPE is strongly dependent on the \emph{calibration} of estimated behaviour policy models: how precisely the behaviour policy is estimated from data. We show how powerful parametric models such as neural networks can result in highly uncalibrated behaviour policy models on a real-world medical dataset, and illustrate how a simple, non-parametric, k-nearest neighbours model produces better calibrated behaviour policy estimates and can be used to obtain superior importance sampling-based OPE estimates.
\end{abstract}

\section{Introduction}

In many decision-making contexts, one wishes to take advantage of already-collected data (for example, website interaction logs, patient trajectories, or robot trajectories) to estimate the value of a novel decision-making policy.  This problem is known as \emph{Off-Policy Policy Evaluation} (OPE), where we seek to determine the performance of an \emph{evaluation} policy, given only data generated by a \emph{behaviour} policy.  Most OPE procedures \citep{precup2000eligibility,off-policy-eval,thomas2016data,mrdr} rely (at least partially) on the technique of Importance Sampling (IS) which, when used in RL, requires the behaviour policy to be known.  However, for observational studies in domains such as healthcare, we do not have access to this information. One way to handle this is to estimate the behaviour policy from the data, and then use it to do importance sampling-based OPE. However, the quality of the resulting OPE estimate is critically dependent on the \emph{calibration} of the behaviour policy -- how precisely it is estimated from the data, and whether the probabilities of actions under the approximate behaviour policy model represent the true probabilities.  

In this work, we evaluate the sensitivity of off-policy evaluation to calibration errors in the learned behaviour policy.  In particular, we perform a series of careful empirical studies demonstrating that: 
\begin{enumerate}[nosep]
    \item Uncalibrated behaviour policy models can result in highly inaccurate OPE in a simple, controlled navigation domain. 
    \item In a real-world sepsis management domain, powerful parametric models such as deep neural networks produce highly uncalibrated probability estimates.
    \item A simple, non-parametric, k-nearest neighbours model is better calibrated than all the other parametric models in our medical domain, and using this as a behaviour policy model results in superior OPE.
\end{enumerate}

\section{Background}
In the reinforcement learning (RL) problem, an agent's interaction with an environment can be represented by a Markov Decision Process (MDP), defined by a tuple $\langle\mathcal{S},\mathcal{A},R,P,P_0,\gamma\rangle$, where $\mathcal{S}$ is the state space, $\mathcal{A}$ is the action space, $R(s,a,s')$ is the reward function, $P(\cdot|s,a)$ is the transition probability distribution, $P_0$ is the initial state distribution, and $\gamma\in[0,1)$ is the discount factor. 
A \emph{policy} is defined as a mapping from states to actions, with $\pi(a|s)$ representing the probability of taking action $a$ in state $s$. 

Let $H := (s_0, a_0, r_{0},\ldots, s_{T-1}, a_{T-1}, r_{T-1}, s_{T})$ be a trajectory generated when following policy $\pi$, and $R(H)=\sum_{t=0}^{T-1}\gamma^tr_t$ be the return of trajectory $H$. 
We can evaluate a policy $\pi$ by considering the expected return over trajectories when following it: 
$V^\pi=\mathbb{E}_{H\sim P^\pi_H}\big[R(H)\big]$. The expectation is taken over the probability distribution of trajectories under policy $\pi$. 
Let the value and action-value functions of a policy $\pi$ at a state $s$ or state-action pair $(s, a)$ be $V^\pi(s)$ and $Q^\pi(s,a)$ respectively. These are defined as the expected return of a trajectory starting at state $s$ or state-action pair $(s,a)$, and then following policy $\pi$. We can write $V^\pi=\mathbb{E}_{s_0\sim P_0}\big(V^\pi(s_0)\big)$.

In off-policy policy evaluation (OPE), we seek to estimate, with low mean squared error (MSE), the value $V^{\pi_{e}}$ of an evaluation policy $\pi_e$ given a set of trajectories $\mathcal{D}=\{H^{(i)}\}_{i=1}^n$ generated independently by following a (distinct) behaviour policy $\pi_b$.

Defining the \emph{importance weight} \citep{precup2000eligibility}, $\rho_t = \prod_{i=0}^{t-1} \frac{\pi_e(a_i^{H}|s_i^{H})}{\pi_b(a_i^{H}|s_i^{H})}$ \footnote{We assume henceforth that for all state-action pairs $(s,a)\in \mathcal{S} \times \mathcal{A}$, if $\pi_b(a|s) = 0$ then $\pi_e(a|s) = 0$. }, we can form the stepwise Weighted Importance Sampling (WIS) estimator of $V^{\pi_{e}}$: ${\hat{V}_{\text{step-WIS}}^{\pi_e} = \sum_{i=1}^n\sum_{t=0}^{T-1}\gamma^t \frac{\rho^{(i)}_{t}}{\sum_{i=1}^n\rho^{(i)}_{t}}\,\, r^{(i)}_t}$. In this work, we consider using the Per-Horizon WIS (PHWIS) estimator, which can handle differing trajectory lengths \citep{doroudi2017importance}, to evaluate medical treatment strategies for sepsis. We also provide results using the Per-Horizon Weighted Doubly Robust (PHWDR) estimator, which incorporates an approximate model of $Q^{\pi_e}(s,a)$ to lower the variance of value estimates \citep{off-policy-eval, thomas2016data}. Further information is in the supplementary material.

\section{Impact of Mis-Calibration: Toy Domain}
\label{sec:synth_domain}
\begin{figure}[!h]
 \centering 
 \centerline{\includegraphics[width=3.5in]{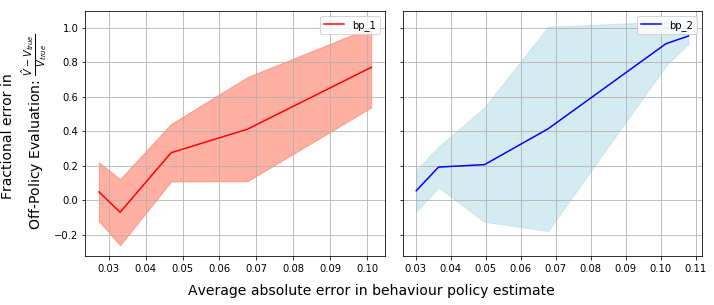} }
 \caption{Mean and standard deviation of the fractional error in OPE, $\frac{\hat{V} - V}{V} $, as a function of the average absolute error in behaviour policy estimation, $\frac{1}{n}\sum_{i=1}^{n}|\pi(a^{(i)}|s^{(i)}) - \hat{\pi}(a^{(i)}|s^{(i)})|$, for two different behaviour policies. The error is calculated with respect to using WIS with the true behaviour policy. The quality of OPE is strongly dependent on the quality of behaviour policy estimation.}
 \label{fig:bp_ope_err} 
\end{figure}

We firstly consider the effect of poorly calibrated behaviour policy models on OPE in a synthetic domain.
The domain is a continuous 2D map ($s \in \mathbb{R}^2$) with a discrete action space, $\mathcal{A} = \{1, 2, 3, 4, 5\}$, with actions representing a movement of one unit in one of the four coordinate directions or staying in the current position. Gaussian noise of zero mean and specifiable variance is added onto the state of the agent after each action. An agent starts in the top left corner of the domain and receives a positive reward within a given radius of the top right corner, and a negative reward within a given radius of the bottom left corner. 
We set the horizon to be 15 in all experiments. A k-Nearest Neighbours (kNN) model is used to estimate the behaviour policy distribution, and its accuracy is varied by adjusting the number of neighbours and training data points used. 
\paragraph{The quality of OPE is strongly dependent on the quality of behaviour policy estimation.} Figure \ref{fig:bp_ope_err} illustrates this via relating the average absolute error in the behaviour policy estimation $\frac{1}{n}\sum_{i=1}^{n}|\pi(a^{(i)}|s^{(i)}) - \hat{\pi}(a^{(i)}|s^{(i)})|$,  to the fractional error in OPE using the WIS estimator, for two different behaviour policies. The error is calculated with respect to using WIS with the true behaviour policy.
Average absolute errors in behaviour policy models of as small as 0.06 can incur errors of up over 50\% in the estimated value --  having a well-calibrated model of the behaviour policy is therefore critical for good OPE. 

\section{Model calibration in the sepsis domain}
As a case-study, we consider the challenge of obtaining well-calibrated behaviour models on a real-world dataset, used in \citet{komorowski} and \citet{raghu2017continuous}, dealing with the medical treatment of sepsis patients in intensive care units (ICUs). 
We use the same framing as \citet{raghu2017continuous}, where the medical treatment process for a sepsis patient is framed as a continuous state-space MDP. A patient's state is represented as a vector of demographic features, vital signs, and lab values. Our state representation concatenates the the previous three timesteps' raw state information to the current time's state vector to capture trends over time.
The action space, $\mathcal{A}$, is of size 25 and is discretised over doses of two drugs commonly given to sepsis patients.
The reward $r_t$ is positive at intermediate timesteps when the patient's wellbeing improves, and negative when it deteriorates. At the terminal timestep of a patient's trajectory, a positive reward is assigned for survival, and a negative reward otherwise.

\subsection{Obtaining well-calibrated behaviour policy models}
We consider modelling the behaviour policy, $\mu(a|s)$ via supervised learning. Importantly, IS uses probabilities (rather than class labels) and hence we require a well-calibrated model, not just an accurate one. To evaluate calibration, we draw a series of test states $s_i$ from a held out test set, and calculate the total variation distance between the predictive distribution over actions from the estimated model, $\hat{\mu}(\cdot|s_i)$, and a ground-truth distribution obtained by considering the empirical distribution over actions from the k-nearest neighbours of the state $s_i$ on the held-out test set, using a custom distance kernel that assesses physiological similarity. Intuitively, states that are physiologically similar should have similar treatment (behaviour policy) distributions. For more information, see the supplementary material. 

\paragraph{Approximate kNN produces better calibrated probabilities than parametric models.} Table \ref{tab:calib} shows the average total variation distance (over 500 test states) between the estimated and target behaviour distributions for different approximate behaviour policy models: logistic regression (LR), random forest (RF), neural network (NN), and an approximate kNN model using random projections \citep{indyk1998approximate} (used instead of full kNN for its computational efficiency). The parametric models are poorly calibrated, especially for sampled states with high severity, where there are fewer data points available for estimation.

\begin{table}[h]
\centering
\begin{tabular}{@{}ccccc@{}}
\toprule
Severity    & LR    & RF    & NN    & Approx kNN     \\ \midrule
0 - 4   & 0.249 & 0.214 & 0.213 & \textbf{0.129} \\ 
5 - 9   & 0.269 & 0.254 & 0.246 & \textbf{0.152} \\ 
10 - 13 & 0.309 & 0.309 & 0.399 & \textbf{0.210} \\ 
14 - 23 & 0.356 & 0.337 & 0.426 & \textbf{0.199} \\ \bottomrule
\end{tabular}
\caption{Average total variation distance between the estimated and target behaviour policy distributions for different models; logistic regression (LR), random forest (RF), neural network (NN), and approximate kNN; stratified by patient severity score (SOFA). The parametric models are surprisingly uncalibrated. Approximate k-nearest neighbours has the best results.}
\label{tab:calib}
\end{table}

\begin{figure}[h]
\centering     %%% not \center
\subfloat[Overconfident predictions]{\label{fig:calib_overconf}\includegraphics[width=80mm]{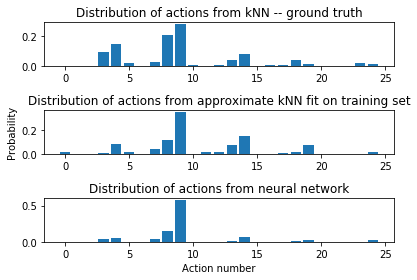}}
\newline
\subfloat[Incorrect predictions]{\label{fig:calib_incorr}\includegraphics[width=80mm]{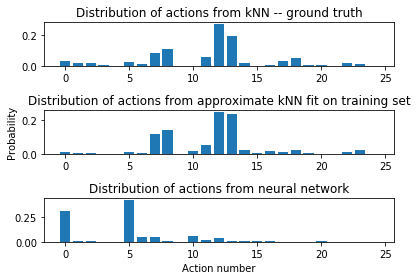}}
\caption{Examples of how neural networks can suffer from poor calibration as behaviour policy models, via overconfident (Figure \ref{fig:calib_overconf}) and incorrect predictions (Figure \ref{fig:calib_incorr}). The approximate kNN model does not suffer from these issues.}
\label{fig:calib_err}
\end{figure}

\paragraph{Neural networks can produce overconfident and incorrect  probability estimates.} Figure \ref{fig:calib_err} shows example predictive distributions over actions for the neural network and approximate kNN as compared to the ground truth, demonstrating over-confident predictions (a result noted by \citet{guo2017calibration}) and incorrect predictions produced by the neural network. Approx kNN may therefore be more appropriate as a behaviour policy model for OPE.

\section{OPE in the sepsis domain}
We now use these behaviour policy models for OPE in the sepsis domain. 
To obtain ground truth for evaluation, we divide our dataset into two subsets $\mathcal{D}_1$ and $\mathcal{D}_2$. 
We can use the behaviour policy from $\mathcal{D}_1$, $\pi_1$, as the evaluation policy with $\mathcal{D}_2$. As we have trajectories with $\pi_1$ as the behaviour policy in $\mathcal{D}_1$, we can average returns on these trajectories to get an on-policy estimate of $V^{\pi_1}$. Low mean squared error between the OPE estimate and the on-policy estimate provides an indication of correctness. 

Two methods of splitting the trajectories are considered: \emph{random} and \emph{intervention} splitting.
In \emph{random} splitting, we randomly select half the trajectories to go in one set, and half to go in the other. 
In \emph{intervention} splitting, the evaluation set contains half of the patients who were never treated with vasopressors (chosen randomly from all such patients), and the training set contains the remainder of patients.
For both methods, results are averaged over different behaviour/evaluation policy pairs -- 50 for PHWIS and 10 for PHWDR.

In the limit of infinite data, random splitting results in identical behaviour and evaluation policies. In our setting, with limited data, the two policies are close (average total variation distance $\approx 0.09$) but this splitting method still permits basic assessment of OPE quality. The average total variation distance with intervention splitting is approximately 0.29.

We estimate $\textnormal{MSE}(V^{\pi_1},\hat{V}^{\pi_1})$ using a bootstrapped method: 
\begin{enumerate}[nosep]
\item Sample $n = 200$ trajectories from $\mathcal{D}_2$.
\item Obtain $\hat{V}^{\pi_1}$ via an OPE method. 
\item Repeat this process $k = 500$ times, representing samples from the distribution of $\hat{V}^{\pi_1}$.
\item Compute the MSE between these samples and ${V}^{\pi_1}$.
\end{enumerate}

\paragraph{The approximate kNN behaviour policy model often results in the best OPE.} Table \ref{tab:ope_res} presents the MSE when using the PHWIS and PHWDR estimators for OPE. 
The estimate for $Q^{\pi_e}(s,a)$ in the PHWDR estimator was obtained using Fitted-Q Iteration (FQI) with random forests \citep{ernst2005tree}.
When using the PHWIS estimator, approximate kNN gives appreciably lower MSE than the neural network (NN), reinforcing the idea that it is better calibrated models can result in better OPE.
The results with the PHWDR estimator do not show as clear a dependence on the behaviour policy. This is because the Approximate Model (AM) terms in one case (random splitting) give low MSE estimates (MSE = 0.177),  and in the other case (intervention splitting) give high MSE estimates (MSE = 3.87). There is therefore less of a dependence on the behaviour policy; OPE is dominated by the AM terms.

\begin{table}[h]
\centering
\begin{tabular}{@{}lcc@{}}
\toprule
                                              & Approx kNN    & NN \\ \midrule
Random split, $\hat{V}^{\pi_e}_{\text{PHWIS}}$        & \textbf{2.48} & 4.04           \\
Intervention split, $\hat{V}^{\pi_e}_{\text{PHWIS}}$ & \textbf{2.04} & 4.65           \\
Random split, $\hat{V}^{\pi_e}_{\text{PHWDR}}$       & 2.04          & \textbf{2.02}  \\
Intervention split, $\hat{V}^{\pi_e}_{\text{PHWDR}}$ & \textbf{3.90} & \textbf{3.90}           \\ \bottomrule
\end{tabular}
\caption{Average MSE when using PHWIS and PHWDR to evaluate different behaviour policies (from random and intervention splitting) in the medical domain with different behaviour policy models: approximate kNN and neural network (NN). The approximate kNN behaviour policy model results in better OPE with PHWIS (pure importance sampling estimator). Approximate model (AM) terms in the PHWDR estimator make the dependence on behaviour policy less clear as the AM terms dominate.}
\label{tab:ope_res}
\end{table}

\section{Conclusion}
\label{sec:conclusion}
In this work, we considered the problem of behaviour policy estimation for Off-Policy Policy Evaluation (OPE), focusing an application in healthcare -- evaluating medical treatment strategies for patients with sepsis. Via a series of empirical studies, we showed how well-calibrated behaviour policy models are highly important for good-quality OPE, and powerful parametric models such as neural networks can often give uncalibrated probability estimates. We demonstrated that a simple, non-parametric, k-nearest neighbours (kNN) behaviour policy model has better calibration than parametric models and that using this kNN model for OPE led to improved results in this real world domain. 
The proposed procedure can be used in other situations where the behaviour policy is unknown, and could improve the quality of OPE estimates, which is an important step towards the use of reinforcement learning in real-world domains.

\section{Acknowledgements} 
This work was supported in part by the Harvard Data Science Initiative, Siemens, and a NSF CAREER grant.

\bibliography{example_paper}
\bibliographystyle{icml2018}

\input{appendix.tex}

\end{document}

%% file: appendix.tex
\onecolumn
\section*{A  \quad Off-Policy Policy Evaluation estimators}
In off-policy policy evaluation (OPE), we consider the situation where we would like to estimate the value $V^{\pi_{e}}$ of an evaluation policy $\pi_e$ given a set of trajectories $\mathcal{D}=\{H^{(i)}\}_{i=1}^n$ generated independently by following a (distinct) behaviour policy $\pi_b$. 
We would like the estimator $\hat{V}^{\pi_e}$ to have low mean squared error (MSE), defined as follows:
${\textnormal{MSE}(V^{\pi_e},\hat{V}^{\pi_e})={\mathbb{E}_{P^{\pi_b}_H}\big((V^{\pi_e}-\hat{V}^{\pi_e})^2\big)}}$. Note that when we have trajectories from $\pi_e$, we can form an estimate of $V^{\pi_e}$ using ${\hat{V}^{\pi_e} = \frac{1}{N}\sum_{i=1}^N R(H_i)}$, which is the Monte-Carlo estimator.

Let us define the quantity $\rho_t = \prod_{i=0}^t \frac{\pi_e(a_i^{H}|s_i^{H})}{\pi_b(a_i^{H}|s_i^{H})}$. This is the \emph{importance weight} \citep{precup2000eligibility}, and is equal the ratio of the probability of the first $t+1$ steps of trajectory $H$ under $\pi_e$ to the probability under $\pi_b$. \footnote{We assume henceforth that for all state-action pairs $(s,a)\in \mathcal{S} \times \mathcal{A}$, if $\pi_b(s|a) = 0$ then $\pi_e(s|a) = 0$. }. Using this definition, we can form the importance sampling estimator of  $V^{\pi_e}$: $\hat{V}_{\text{IS}}^{\pi_e}=\frac{1}{n}\sum_{i=1}^n \rho^{(i)}_{T-1}\sum_{t=0}^{T-1}\gamma^t  \,\,r^{(i)}_t$

Let us also define $\hat{V}_M^\pi(s)$ and $\hat{Q}_M^\pi(s,a)$ to be estimates of the state and action value functions for policy $\pi$ respectively under the approximate model (AM) of the MDP, $M$. We can use an approximate model $M$ to directly find $V^{\pi_e}$. For example, we can write: \newline $\hat{V}_{\text{AM}}^{\pi_e} = \frac{1}{n}\sum_{i=1}^n\sum_{a\in \mathcal{A}}\pi_e(a|s^{(i)}_0)\hat Q^{\pi_e}_M(s^{(i)}_0,a)$.

To estimate the quantity $V^{\pi_e}$, prior work has mainly used one or both of the techniques of Importance Sampling (IS) and Approximate Model (AM) estimation \citep{thomas2016data}. 
The IS approach to evaluation relies on using the importance weights $\rho_t$ to adjust for the difference between the probability of a trajectory $H$ under the behaviour policy $\pi_b$ and the probability under the evaluation policy $\pi_e$. 
Two commonly used estimators in the IS family, which improve on the simple IS estimator are the step-wise IS and step-wise WIS estimators, defined as follows (with $i$ indexing the trajectories in $\mathcal{D}$): 
\begin{equation*}
\begin{split}
\hat{V}_{\text{step-IS}}^{\pi_e} =\frac{1}{n}\sum_{i=1}^n\sum_{t=0}^{T-1}\gamma^t \rho^{(i)}_{t} \,\,r^{(i)}_t \qquad
\hat{V}_{\text{step-WIS}}^{\pi_e} = \sum_{i=1}^n\sum_{t=0}^{T-1}\gamma^t \frac{\rho^{(i)}_{t}}{\sum_{i=1}^n\rho^{(i)}_{t}}\,\, r^{(i)}_t
\end{split}
\end{equation*}
The step-IS estimator is an unbiased estimator of $V^{\pi_e}$ but suffers from high variance (due to the product of importance weights). 
The step-WIS estimator is biased, but consistent, and has lower variance than step-IS. However, its variance can often still be unacceptably high \citep{thomas2016data}. 
These IS estimators can have significant bias when the behaviour policy is unknown.

In AM estimation, we use the approximate model $M$ to directly find $V^{\pi_e}$, as defined earlier. It may be difficult to trust these estimators, however, given that we cannot always find their bias and variance.

Doubly Robust methods \citep{off-policy-eval,thomas2016data} combine IS and AM techniques together in order to reduce the variance of the resulting estimator. The Weighted Doubly Robust (WDR) estimator, which has demonstrated effective empirical performance \citep{thomas2016data}, is defined as follows, with $w_t^{(i)} = \frac{\rho^{(i)}_{t}}{\sum_{i=1}^n\rho^{(i)}_{t}}$:

\begin{equation*}
\begin{split}
\hat{V}_{\text{WDR}}^{\pi_e} = \frac{1}{n} \sum_{i=1}^n\sum_{t=0}^{T-1}\Big(\gamma^tw_{t}^{(i)}r^{(i)}_t - 
 \gamma^t \, \big(w_{t}^{(i)} \, \hat{Q}_M^{\pi_e}(s^{(i)}_t,a^{(i)}_t)-w_{t-1}^{(i)} \, \hat{V}_M^{\pi_e}(s^{(i)}_t)\big)\Big)
\end{split}
\end{equation*}

Note that these estimators are valid for trajectories with the same length; extensions to handle trajectories of different length can be found in \citet{doroudi2017importance} -- this is called the \emph{Per-Horizon} extension (resulting in the PHIS and PHWIS estimators).

\citet{doroudi2017importance} defined the Per-Horizon Weighted Importance Sampling (PHWIS) estimator as follows:

$$\hat{V}_{\text{PHWIS}}^{\pi_e}= \sum_{l \in \mathcal{L}} W_l \, \,
\frac{1}{\sum_{\{\tau_i | T_i = l\}} \rho_{T_i-1}^{(i)}} \,
\sum_{\{\tau_i | T_i = l\}}\rho_{T_i-1}^{(i)}\sum_{t=0}^{T_i-1}\gamma^t  \,\,r^{(i)}_t $$
where $\mathcal{L}$ is the set of all trajectory lengths, and $W_l$ is the fraction of the total number of trajectories $n$ with length equal to $l$:
$W_l = \frac{|\, \{\tau_i | T_i = l\} \, |}{n}$

% \begin{equation*}
% \begin{split}
% \hat{V}_{\text{PHWIS}}^{\pi_e} = \sum_{l \in \mathcal{L}} W_l \, \,
% \frac{1}{\sum_{\{\tau_i | T_i - 1 = l\}} \rho_{T_i-1}^{(i)}}  
% \sum_{\{\tau_i | T_i - 1 = l\}}\rho_{T_i-1}^{(i)}\sum_{t=0}^{T_i-1}\gamma^t  \,\,r^{(i)}_t
% \end{split}
% \end{equation*}

% where $\mathcal{L}$ is the set of all trajectory lengths, and ${W_l = \frac{|\, \{\tau_i | T_i - 1 = l\} \, |}{n}}$ defining the fraction of the total number of trajectories $n$ with length equal to $l+1$.

This estimator has high variance; we can define a lower variance equivalent by considering a step-wise version:

$$\hat{V}_{\text{step-PHWIS}}^{\pi_e}= \sum_{l \in \mathcal{L}} W_l \, \,
\sum_{\{\tau_i | T_i = l\}}\sum_{t=0}^{T_i-1} \frac{\rho_{t}^{(i)}}{\sum_{\{\tau_i | T_i = l\}} \rho_{t}^{(i)}} \, \gamma^t  \,\,r^{(i)}_t $$

We can also introduce control variates into the estimator and form the Per-Horizon Weighted Doubly Robust (PHWDR) estimator, as follows. First, let us define $\hat{V}_{\text{WDR}, \, l}^{\pi_e}$ to be the WDR estimator given all trajectories of length $l$. We can write this as follows, with $w_{t,l}^{(i)} = \frac{\rho_{t}^{(i)}}{\sum_{\{\tau_i | T_i = l\}} \rho_{t}^{(i)}}$:

\begin{equation*}
\begin{split}
\hat{V}_{\text{WDR}, \, l}^{\pi_e} = \sum_{\{\tau_i | T_i = l\}}\sum_{t=0}^{T-1}\Big(\gamma^tw_{t,l}^{(i)}r^{(i)}_t - 
 \gamma^t \, \big(w_{t,l}^{(i)} \, \hat{Q}_M^{\pi_e}(s^{(i)}_t,a^{(i)}_t)-w_{t-1,l}^{(i)} \, \hat{V}_M^{\pi_e}(s^{(i)}_t)\big)\Big)
\end{split}
\end{equation*}

Then, it is straightforward to write, with $W_l$ as defined before: 
$$\hat{V}_{\text{PHWDR}}^{\pi_e}= \sum_{l \in \mathcal{L}} W_l \, \, \hat{V}_{\text{WDR}, \, l}^{\pi_e} $$

\section*{B \quad Assessing Behaviour Policy Calibration}
To evaluate the calibration of models, we can calculate the distance between the estimated behaviour policy and target behaviour policy. 
In order to calculate this distance, we require the target behaviour policy, which is unknown. 
However, we can use domain knowledge to inform the choice of the target distribution.  
In this medical setting, we propose that what governs the clinician's choice of action is the physiological state of the patient, and that patients with {\em similar} physiological states will be treated in similar ways. This is a reasonable approximation, given that the state encodes the patient's physiology effectively \citep{raghu2017continuous}. 

We define similarity of patient states using a `physiological distance kernel', which is based on Euclidean distance and upweights certain informative features of the patient's state.  Informative features were the patient's SOFA score, lactate levels, fluid output, mean and diastolic blood pressure, Pa$\text{O}_2$/Fi$\text{O}_2$ ratio, chloride levels, weight, and age. These are clinically interpretable: the SOFA score and lactate levels provide indications of sepsis severity; careful monitoring of a patient's fluid levels is essential when managing sepsis \citep{marik2017fluid}; and blood pressure indicates whether a patient is in septic shock. These features are upweighted by a factor of 2 in our distance kernel (where $D = 198$, the dimensionality of our state representation):

$$
k(\mathbf{s},\mathbf{s}')= \sum_{i=1}^D w_i (s_i - s_i')^2 
\begin{cases}
w_i = 2 , \text{ for \emph{informative} } i \\
w_i = 1 , \text{ otherwise}\\
\end{cases}
$$

To find the target distribution for a given test state, we use a k-nearest neighbour (kNN) estimate with this distance kernel and form an empirical distribution of the actions taken from the test set neighbours. We consider 150 neighbours to provide reasonable coverage in the estimate. A Ball Tree data structure is used for efficiency. Querying this data structure is computationally expensive ($\sim$1 second per query), so we sample 500 states for patients at different severities (range of SOFA score) and average results for these sets. We use the total variation distance, defined as $\delta(\pi_b(\cdot|s), \hat{\pi}_b(\cdot|s)) = \frac{1}{2}\sum_{a \in \mathcal{A}}|\pi_b(a|s) - \hat{\pi}_b(a|s)|$ for the discrete action space, as the distance metric. Our approximate behaviour policies are trained on a separate training dataset and we compare the predictive distribution over actions for the test states to the result from the kNN estimate.